\pdfoutput=1

\documentclass[11pt]{article}

\usepackage{emnlp2021}

\usepackage{times}
\usepackage{latexsym}

\usepackage[T1]{fontenc}

\usepackage[utf8]{inputenc}

\usepackage{microtype}
\usepackage{multirow}
\usepackage{array}
\usepackage{subfigure}
\usepackage{graphicx}
\graphicspath{ {images/} }
\usepackage{amsfonts}
\usepackage{mathrsfs}
\usepackage{amsmath}
\makeatletter
\renewcommand{\maketag@@@}[1]{\hbox{\m@th\normalsize\normalfont#1}}%
\makeatother
\usepackage{arydshln}
\usepackage{booktabs}
\usepackage{verbatim}
\usepackage{algorithm}  
\usepackage{algpseudocode}  
\usepackage[symbol]{footmisc} 

\usepackage{float}
\usepackage{hyperref}

\makeatletter
\def\UrlAlphabet{%
      \do\a\do\b\do\c\do\d\do\e\do\f\do\g\do\h\do\i\do\j%
      \do\k\do\l\do\m\do\n\do\o\do\p\do\q\do\r\do\s\do\t%
      \do\u\do\v\do\w\do\x\do\y\do\z\do\A\do\B\do\C\do\D%
      \do\E\do\F\do\G\do\H\do\I\do\J\do\K\do\L\do\M\do\N%
      \do\O\do\P\do\Q\do\R\do\S\do\T\do\U\do\V\do\W\do\X%
      \do\Y\do\Z}
\def\UrlDigits{\do\1\do\2\do\3\do\4\do\5\do\6\do\7\do\8\do\9\do\0}
\g@addto@macro{\UrlBreaks}{\UrlOrds}
\g@addto@macro{\UrlBreaks}{\UrlAlphabet}
\g@addto@macro{\UrlBreaks}{\UrlDigits}
\makeatother
\title{Less is Better: Recovering Intended-Feature Subspace\\ to Robustify NLU Models }

\author{Ting Wu \\
   Fudan Univerisity\\
  \texttt{tingwu21@m.fudan.edu.cn} \And
  Tao Gui \footnotemark\\
  Fudan Univerisity\\
  \texttt{tgui@fudan.edu.cn} \\
  }

\begin{document}
\maketitle
\footnotetext[1]{Corresponding author.}
\begin{abstract}

\renewcommand{\thefootnote}{\arabic{footnote}}
Datasets with significant proportions of \textit{bias} present threats for training a trustworthy model on NLU tasks. Despite yielding great progress, current debiasing methods impose excessive reliance on the knowledge of bias attributes. Definition of the attributes, however, is elusive and varies across different datasets.
Furthermore, leveraging these attributes at input level to bias mitigation may leave a gap between intrinsic properties and the underlying decision rule. To narrow down this gap and liberate the supervision on bias, we suggest extending bias mitigation into feature space. Therefore, a novel model, \textbf{R}ecovering \textbf{I}ntended-Feature \textbf{S}ubspace with \textbf{K}nowledge-Free (RISK) is developed. Assuming that shortcut features caused by various biases are unintended for prediction, RISK views them as redundant features. When delving into a lower manifold to remove redundancies, RISK reveals that an extremely low-dimensional subspace with \textit{intended features} can robustly represent the highly biased dataset. Empirical results demonstrate our model can consistently improve model generalization to out-of-distribution set, and achieves a new state-of-the-art performance~\footnote{Our code and data are available at \url{https://github.com/CuteyThyme/RISK.git}.}.  


\end{abstract}

\section{Introduction}
  




Pretrained language models have achieved remarkable performance on a wide range of natural language understanding (NLU) benchmarks~\cite{devlin-etal-2019-bert}. However, when encountering more challenging test sets, they dramatically fail~\cite{mccoy-etal-2019-right}. Studies indicate such a dilemma is mainly rooted in the model's reliance on specific \textit{dataset biases}~\cite{Gururangan2018AnnotationAI,zhang-etal-2019-selection,schuster-etal-2019-towards}, which correlate well with labels but not for the intended underlying task. For instance, on the natural language inference (NLI) task, models tend to use negation cues ("not", "no", etc.), for a \textit{Contradiction} prediction, whereas a learner intended to learn the \textit{underlying correlation} based on the context semantics.

To train a NLU model that captures the \textit{underlying correlation} from biased datasets, current approaches focus on how to leverage kinds of supervision effectively. One of the most popular forms of such supervision is to explicitly construct a bias-only model under human annotations, e.g., a hypothesis-only model for NLI task, and factor it out from the main model through ensemble-based training~\cite{clark-etal-2019-dont,aclUtamaMG20}.  Another empirical line of research shifts supervision from bias type annotations to weak model learners. They find models with limited capacity~\cite{clark-etal-2020-learning,sanh2021learning} or training on limited dataset~\cite{utama-etal-2020-towards} prone to extract shortcut patterns first, the observation of which can be utilized to mitigate dataset bias. 
\begin{figure}[!t]
\centering
    \includegraphics[scale=0.18]{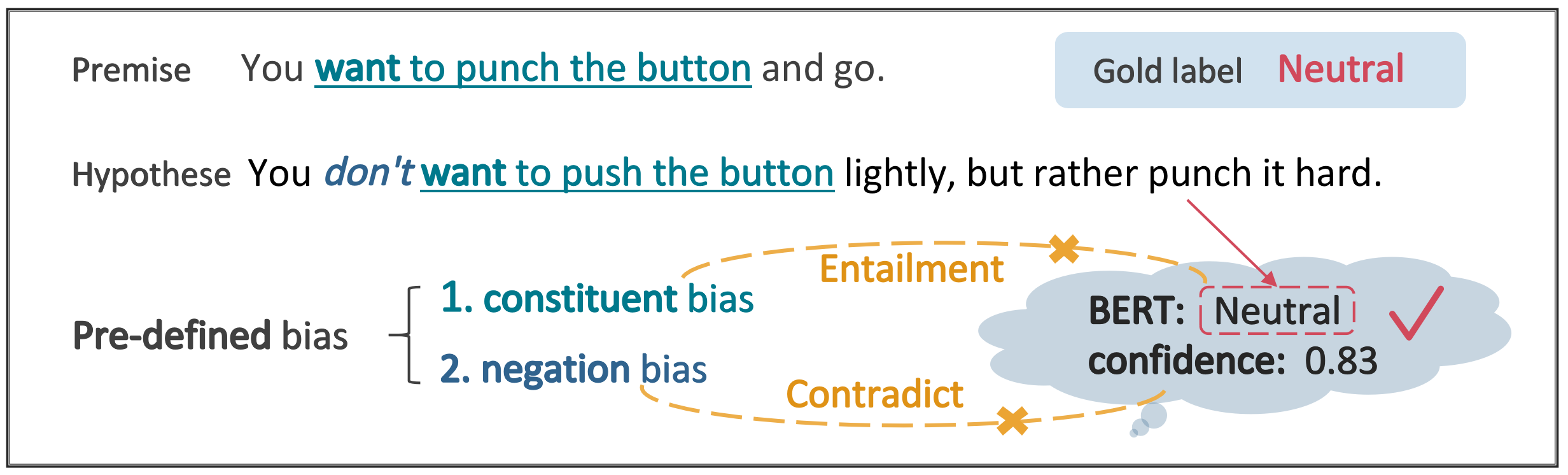}
    \caption{A toy example that illustrate bias in MNLI-matched dev set. BERT's prediction \textit{Neutral} does not comply with the assumed decision rule (Entailment, Contradiction) caused by pre-defined bias. }
    \label{fig:hardness}
\end{figure}

Despite the supervision on bias has shown effectiveness in bias mitigation, the fundamental questions remain unsolved. On one hand, acquirement of supervision on the bias either from human knowledge or model learning behaviours, is often a laborious and expensive cost. Moreover, considering the definition of bias attributes is elusive and varies across datasets, the external knowledge can not cover all types of biases in the dataset, leaving potential bias underexplored, e.g., bias beyond the definition or bias harder to learn. On the other hand, capturing bias only at the input examples is just like a black box, being oblivious to the intrinsic properties that drives model to make prediction. The toy example shown in Figure~\ref{fig:hardness} reveals that predefined bias does not necessarily lead the model to learn the unintended decision rule (i.e., constituent bias triggers an Entailment prediction, negation bias triggers an Contradiction prediction). Hence, current debiasing methods inevitably fall short in above two limitations.  

On account of the consensus that shortcut features induced by biased examples are detrimental for prediction, various kinds of biases can thus be equivalently viewed as redundancies. When delving into feature space, closer to the decision rule to remove these redundant features, supervision of the attributes from biased examples can be liberated as well. Therefore, we develop a novel model, \textbf{R}ecovering \textbf{I}ntended-Feature \textbf{S}ubspace with \textbf{K}nowledge-Free (RISK). Aimed with purifying redundancies from feature space, RISK reveals that for a highly biased dataset, a small subset of \textit{informative} and \textit{shared} features, i.e. intended ones, can give rise to a robust prediction. Concretely, RISK maps features into a lower manifold and learns an orthogonal projector spanned by \textit{geometric median subspace} to recover the intended-feature subspace in an end-to-end manner.


Experimental results on three NLU tasks show RISK outperforms other methods by a large margin, indicating its potential to mitigate bias and the prerequisites of supervision on biased attributes can be liberated. Moreover, when transferring to more challenging out-of-distribution set, RISK can consistently improve the robustness of NLU models. To sum up, our contributions are three-fold as follows:
 
 $\bullet$ We propose a novel \textit{feature-based} debiasing model, termed as RISK. RISK is the initial attempt that free of the supervision on bias attributes.

 $\bullet$ We reveal shortcut features as part of redundancy, and thus only leveraging the informative features shared across biased and bias-free examples can achieve the goal of bias mitigation. 

 $\bullet$ We conduct extensive experiments to validate the effectiveness of RISK in mitigating bias. Moreover, RISK exhibits great power to generalize to more challenging scenarios, showing its potential to robustify NLU models.

\section{Bias Mitigation As Feature Redundancy}
\subsection{Problem Setup}
Given training dataset $\mathcal{D}=\{x_i, y_i\}_{i=1}^N$ including $C$ classes, a NLU task requires the model to understand the semantic of input text $x_i$ and then predict the target label $y_i$. Generally, the model is composed of a feature extractor $\mathcal{F}(\cdot): \mathcal{F}(\mathbf{x}) \rightarrow \mathbf{z}$ and a linear classifier $g(\cdot): g(\mathbf{z}) \rightarrow \mathbf{\hat{y}}$.   

When the model is trained on a highly biased dataset, it will easily capture shortcut features in high-dimensional $\mathbf{z}$. Since the shortcut features are the unintended ones that induce predictions, we treat them as a kind of redundancy. Therefore, mitigating dataset bias can be subsumed under minimizing redundancy in feature space. 


\subsection{Feature Redundancy by Subspace Modeling}
In statistical machine learning, feature subspace paves a path towards eliminating redundant features, as it sheds light on projecting high-dimensional feature onto one subspace, which can significantly capture its most significant information.
A common formulation for subspace modeling is to find an orthogonal projection $\mathbf{P}$ of dimension $d$ whose subspace can robustly represents the input features~\cite{robustsubspace}. Let $\mathbf{I}$ denote the identity matrix in the ambient space of the high-dimensional feature $\mathbf{z}$, and the least $q$-th power deviations formulation for $q>0$ seeks $\mathbf{P}$ that minimizes:
\begin{equation}
    \label{equation:energy}
    \mathcal{L(\mathbf{P})} = \sum_{i=1}^N\big\Vert(
    \mathbf{I}-\mathbf{P})z_i\big\Vert_2^q 
\end{equation}

Classically, taking $q=2$ results in principal component analysis(PCA), which finds the orthogonal directions of maximum variance:

\begin{equation}
    \mathcal{L(\mathbf{P})} = \sum_{i=1}^N\big\Vert(
    \mathbf{I}-\mathbf{P})z_i\big\Vert_2^2  \nonumber
\end{equation}

\begin{figure*}[!t]
    \centering
    \includegraphics[scale=0.31]{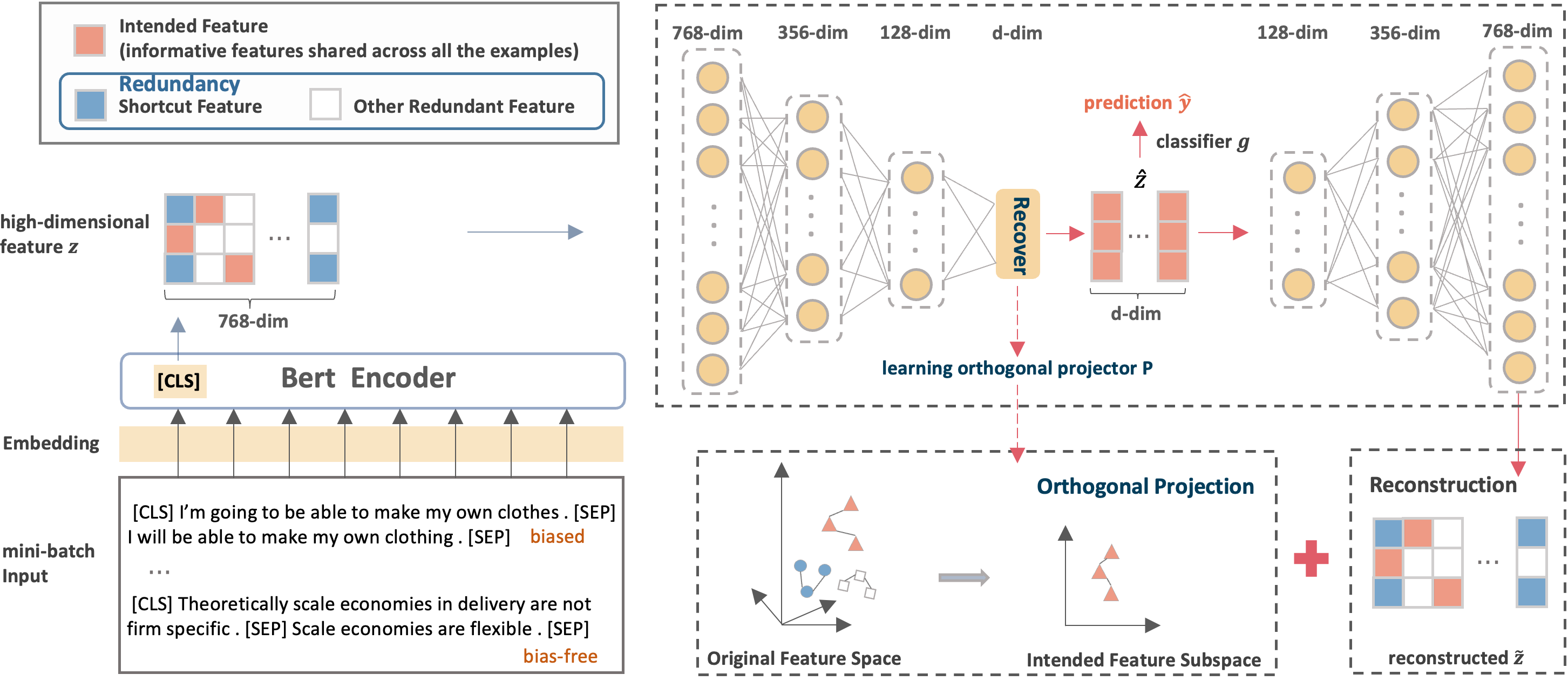}
    \caption{Model Architecture of RISK.}
    \label{fig:main_model} 
\end{figure*} 

\subsection{Geometric Median Subspace as solution.}\label{sec:geometric}
However, even approximate minimization of Eq.~\ref{equation:energy} is nontrivial, since it has been shown to be NP hard for $1\leq q<2$, furthermore, $q<1$ can result in a wealth of local minima. Literature have theoretically proven the preferable minimization is $q=1$~\cite{osborne1985analysis, nyquist1988least}, and thus equals to replace the least squares formulation in PCA with least absolute deviations as follows:
\begin{equation}
\mathcal{L(\mathbf{P})} = \sum_{i=1}^N\big\Vert (\boldsymbol{I}-\boldsymbol{P})z_i \big\Vert_2^1
\label{eq: median_subspace}
\end{equation}

A nice interpretation of the minimizer of above equation is a \textit{Geometric Median Subspace}~\cite{Fletcher2009TheGM}, analogous geometric median in modeling centers of input features. Ideally, once we solve a orthogonal projection spanned by this geometric median space, we can achieve an robust estimation of all input features. Since the shortcut features are not shared by bias-free examples, they will be removed automatically as redundancy.

\section{RISK: Feature-based Debiasing Without Supervision on Bias}\label{sec:method}
Guided by the theoretical subspace modeling discussed above, in this section, we illustrate the detailed implementation of RISK. In practice, we adopt autoencoder as the main architecture. Leveraging autoencoder to map features into a lower manifold is the first stage of removing redundant features. We further add a simple but effective Recovery Layer within autoencoder to learn a orthogonal projection $\mathbf{P}$, leaving the shared informative features to perform final predictions.

\subsection{Delving into Feature Space}
 We use BERT $\mathcal{F}_\theta$ to map each textual data point $x_i$ into a high-dimensional feature space, that is, $\mathbf{z} = \mathcal{F}(\mathbf{x}, \theta)$. To be specific, $\mathbf{z}$ corresponds to [CLS] token embedding the last layer BERT outputs. It has been convinced that embeddings from pre-trained language models contain much redundency for down-stream NLU tasks~\cite{dalvi-etal-2020-analyzing}. As for highly biased dataset, $\mathbf{z}$ will easily capture substantial shortcut ones. We thus categorize $\mathbf{z}$ into following two feature types:

\textbf{Intended Features} are the \textit{informative} features \textit{shared} across biased and bias-free examples.

\textbf{Redundant Features} include shortcut features that only correlate well with labels, and other redundant features (e.g., task-irrelevant, task-relevant but non-robust ones).

\subsection{Autoencoder: To Be Informative Features}
 For the first stage of mitigating feature redundancy as a way of bias mitigation, we opt to employ an encoder $\mathscr{E}$ composed of a three-layer MLP to map $\mathbf{z}$ into a lower manifold.

\textbf{Reconstruction loss.} As shown in Figure~\ref{fig:main_model}, to be symmetric of the encoder $\mathscr{E}$ that project $\mathbf{z}$ into a lower manifold, we also train a decoder $\mathscr{D}$ that map $\mathbf{\hat{z}}$ into $\mathbf{\tilde{z}}$, i.e., the reconstruction representation of $\mathbf{z}$, formulating a bottleneck autoencoder as result. The reconstruction loss is thus defined as:
\begin{equation}
    \mathcal{L}_{\rm recon} = \sum_{i=1}^N \big\Vert z_i - \mathscr{D}(\hat{z}_i) \big \Vert_2^2
\end{equation}
The reconstruction term is used to ensure that a good reconstruction of the original feature can be obtained by using the learned low-dimensional subspace features. Notably, as we defined, \textit{informative} is one of the key characteristics of the intended features. We can further prove that minimize the reconstruction error can serve as maximizing the lower bound of the mutual information between $\mathbf{z}$ and $\mathbf{\hat{z}}$. In general $\mathbf{\hat{z}}$ is not an exact reconstruction of $\mathbf{z}$, but rather in probabilistic terms as the mean of a distribution $p=(Z|\hat{Z}=\mathbf{\hat{z}})$, this yields an associated reconstruction error~\cite{vincent2010stacked} to be optimized:
\begin{equation}
    \mathcal{L}_{\rm recon} \propto -{\rm log} p(\mathbf{z}|\mathbf{\hat{z}}) \nonumber
\end{equation}
In conjuction with it, minimizing the reconstruction loss actually carry the following optimization:
\begin{equation}
    {\rm min} \mathbb{E}[\mathcal{L}_{\rm recon}(\mathbf{z}, \mathbf{\hat{z}})] = {\rm max} \mathbb{E}[log\mathbb{P}(\mathbf{z}|\mathbf{\hat{z}})]
\end{equation}
Maximizing the expectation of the conditional probabilty $\mathbb{E}[log\mathbb{P}(\mathbf{z}|\mathbf{\hat{z}})]$ is equivalent to maximizing the mutual information between $\mathbf{z}$ and $\mathbf{\hat{z}}$~\cite{chen2022preserving}. This promises the subspace where $\mathbf{\hat{z}}$ lies in is informative and task-relevant for downstream task. 

\subsection{Recovery Layer: To Be Shared Features}
In fact, only utilizing autoencoder can not promise the latent subspace as the intended one we defined before, since shortcut features dominated by biased examples also contain useful but not robust information for prediction. Therefore, going a step further to remove redundant features is needed. 

\textbf{Projection Loss.} Leveraging the core idea of subspace modeling, \textit{geometric median subspace} is a preferable minimum to solve the shared features in ideal. In this way, we can recast the problem into learning an orthogonal projector spanned by such median subspace. With the expansion of Eq. \ref{eq: median_subspace}, the following projection loss function can be achieved:
\begin{equation}
\begin{split}
    \mathcal{L}_{\rm proj}(\mathbf{A}) &= \lambda_1 \sum_{i=1}^N \big\Vert z_i-\mathbf{A}^\top \mathbf{A}z_i\big\Vert_2^1 \\
    &+ \lambda_2 \big\Vert \mathbf{A}\mathbf{A}^\top -\mathbf{I}_d\big\Vert_F^2 
\end{split}
\end{equation} 
we use $\mathbf{A}$ to denote the transformation that reduces feature dimension to $d$, and $\mathbf{A}^\top$ denotes the transpose of $\mathbf{A}$, $\mathbf{I}_d$ denotes the $d\times d$ identity matrix and $\big\Vert \cdot \big\Vert_{\rm F}$ denotes the Frobenius norm. Here $\lambda$ is an hyperparameter represent the weight of the projection loss to the whole learning objective, for the simplicity, we let $\lambda_1=\lambda_2$. We later show it associates with the dataset characteristics in Sec.~\ref{sec:analysis}.   

It can be noted that the first term in the weighted sum of above loss function is close to Equation~\ref{eq: median_subspace} as long as $\mathbf{A}\mathbf{A}^\top$ is close to an orthogonal projector. To enforce this requirement, we introduce the second term that imposes the nearness of $\mathbf{A}\mathbf{A}^\top$ to an orthogonal projection. 

Practically, the transformation $\mathbf{A}$ is implemented as a linear MLP layer within the autoencoder $\mathscr{E}$, coined as the Recovery Layer. By applying the projection loss, the parameters of the trained Recovery Layer can approximate the minimal result of Eq.~\ref{eq: median_subspace}. In a sense, the Recovery Layer can be considered as bridging the connections between statistical machine learning and DNN.

\subsection{Predictors Fitting in the Intended-Feature Subspace}
Intuitively, as a robust model to defend against various distribution shift, it is expected to learn an optimal predictor $g$, which relies on only the intended features most relevant to current task to make predictions. So for the final step, we just fit a linear classifier in the recovered subspace:
\begin{equation}
    g(\mathbf{\hat{z}}) = W^\top \mathbf{\hat{z}} + b \nonumber
\end{equation}

Along with minimizing the cross entropy between $g(\mathbf{\hat{z}})$ and $\mathbf{y}$, the final learning objective of RISK is summed into:
\begin{equation}
    \mathcal{L}_{\rm RISK} = \mathcal{L}_{\rm CE} + \mathcal{L}_{\rm recon} +  \mathcal{L}_{\rm proj} \nonumber
\end{equation}

With the dual regularization of reconstruction loss and projection loss, we therefore promise the intended-feature subspace is de facto informative and shared.

\section{Experiments}
In this section, we provide comprehensive analysis on RISK through extensive experiments on three NLU tasks, and compare out-of-distribution as well as in-distribution accuracy of RISK with other debiasing methods to demonstrate its strength.

\subsection{Tasks and Biased Datasets}
We evaluate our approach on three NLU tasks: natural language inference (NLI), fact verification, and paraphrase identification.

\textbf{Natural Language Inference} aims to determine whether a premise sentence entails a hypothesis sentence. We use the MNLI dataset~\cite{williams-etal-2018-broad} for training, nevertheless, recent studies indicate that models trained on these NLI datasets tend to adopt shallow heuristics(e.g., lexical overlap, hypothesis-only) to predict~\cite{Gururangan2018AnnotationAI, hypothsesis-only-nli-baselines}. Based on the findings, HANS(Heuristic Analysis for NLI Systems, \citet{mccoy-etal-2019-right}) is designed to contain many examples where the heuristics fail, and we condider it as the challenging set for evaluation.

\textbf{Fact Verification} requires models to validate a claim in the context of evidence. For this task, we use the training dataset provided by the FEVER challenge~\cite{thorne-etal-2018-fever}. Studies show that models ignoring evidence can still achieve high accuracy on FEVER, accordingly, Fever-Symmetric dataset~\cite{schuster-etal-2019-towards} is used as the test sets for evaluation.

\textbf{Paraphrase Identification} is designed to identify whether a pair of sentences have the same thing. We train the models on QQP~\cite{qqpShankar}, a widely used dataset for the task. Similarly to MNLI, models trained on QQP are inclined to mark any sentence pairs with high word overlap as paraphrases despite clear clashes in meaning. As for the balance with respect to the lexical overlap heuristic in PAWS(Paraphrase Adversaries from Word Scrambling, \citet{paws2019naacl}) , we use it as our out-of-distribution set.

\begin{table*}[ht]
\centering
\begin{tabular}{p{4.2cm}ccc|ccc|ccc}
\hline
\multirow{2}{*}{\centering \textbf{Model}} & \multicolumn{3}{c|}{\textbf{MNLI}} &\multicolumn{3}{c|}{\textbf{FEVER}} & \multicolumn{3}{c}{\textbf{QQP}} \\ 
 & ID  & HANS  & $\Delta$  & ID & Symm. & $\Delta$  & ID & PAWS & $\Delta$ \\
\hline
BERT-base & 84.5  & 61.2  & -  & 85.6  & 55.1  & -  & 90.8 & 36.1 & -  \\ \hline

Reweighting & 83.5 & 69.2 & +8.0 & 84.6   & 61.7  & +6.6  & 89.5   & 48.6  & +12.5 \\

Product-of-Experts & 84.1 & 66.3 & +5.1 & 82.3 & 62.0 & +6.9 & 86.9  & 56.5  & +20.4 \\

Learned-Mixin   & 84.2  & 64.0 & +2.8 & 83.3  & 60.4  & +5.3 & 87.6 & 55.7 & +19.6 \\ 

Conf-reg  &  83.4  & 69.1  & +7.9 & 86.4  & 60.5 & +5.4 & 89.1 & 40.0  & +3.9  \\ \hline

Conf-reg$_{\rm \mathbf{self-debias}}$ $\spadesuit$ &  84.3 & 67.1 & +5.9 & 87.6  & 60.2 & +5.1 & 89.0 & 43.0  & +6.9 \\

Weak Learner & 83.3  & 67.9  & +6.7  & 85.3  & 58.5 & +3.4 & - & - & - \\

BERT+$\mathcal{F}_{\rm B \scriptscriptstyle{I} \scriptstyle{LSTM}}$ & 82.9 & 70.4 & +9.2 & 86.5 & 61.7 & +6.6 & 88.0 & 47.6 & +11.5 \\
\hdashline
\textbf{RISK} & \textbf{84.5}  & \textbf{71.3} & \textbf{+10.1} & \textbf{88.3} & \textbf{63.9} & \textbf{+8.8} & \textbf{90.1} & \textbf{56.5} & \textbf{+20.4}\\

\quad w/o Reconstruction Loss & 84.2 & 69.2 & +8.0 & 87.6 & 60.1 & +5.0 & 90.5 & 50.6 & +14.5\\
\quad w/o Projection Loss & 83.9 & 64.6 & +3.4 & 86.5 & 57.7 & +2.6 & 90.4 & 42.1 & +6.0\\
\hline
\end{tabular}
\caption{Model performance(accu.($\%$)) on in-distribution and corresponding challenge test set. $\spadesuit$: Self-debiasing framework is implemented in conjunction with the bias-known-prior models, we select the version that achieves the best performance in the original paper, i.e., \textit{Confidence Regularization} with annealing mechanism. "w/o Reconstruction Loss" represents RISK is trained without the regularization of reconstruction loss, and "w/o Projection Loss" represents RISK is trained without the regularization of projection loss.}
\label{tab: main_res}
\end{table*}

\subsection{Baseline Methods}
We compare RISK against seven debiasing models either with bias known or unknown. As for the bias-known models, supervision on bias is mainly the bias type, and for the bias-unknown models, supervision comes from a shallow model. For all these baseline methods, we adopt the BERT-base model~\cite{devlin-etal-2019-bert} as the main model.

\textbf{Bias-known-prior Models.} \textbf{\romannumeral 1). Reweighting}~\cite{clark-etal-2019-dont} trains on a weighted version of the data to encourage the main model to focus on examples the bias-only model gets wrong. \textbf{\romannumeral 2). Product-of-Experts}~\cite{clark-etal-2019-dont} forces the main model to focus on learning from examples that are not predicted well by the bias-only model via logit ensembling. \textbf{\romannumeral 3). Learned-Mixin}~\cite{clark-etal-2019-dont} further improves this ensemble-based method by parameterizing the ensembling operation, allowing the main model to learn when to incorporate the output from the bias-only model for the ensembled prediction. \textbf{\romannumeral 4). Conf-reg}~\cite{aclUtamaMG20} presents a novel \textit{confidence regularization} method that encourage the main model to make predictions with lower confidence on examples that contained biased features.

\textbf{Bias-known-free Models.} For this line of research, models can bypass the need of hand-engineered bias-specific structures since a shallow model is utilized to identify biased examples automatically.  \textbf{\romannumeral 5). Self-debiasing}~\cite{utama-etal-2020-towards} observe that BERT-base trained on a small subset of the training dataset can grasp the distribution of biased examples. \textbf{\romannumeral 6). Weak Learner}~\cite{sanh2021learning} view models with limited capacity, i.e. Tiny-BERT~\cite{turc2020wellread}, as the shallow one to obtain biased features. 
\textbf{\romannumeral 7). BERT+$\mathcal{F}_{\rm B \scriptscriptstyle{I} \scriptstyle{LSTM}}$}~\cite{yaghoobzadeh-etal-2021-increasing} employ example forgettting to find minority examples, and robustify the model by fine-tuning twice, first on the full training data and second on the minorities only.

\subsection{Implementation Details}
For each task, we utilize the training configurations that have been proven to work well in previous studies, that is, a learning rate of $5e^{-5}$ for MNLI and $2e^{-5}$ for FEVER and QQP, and choose AdamW as optimizer with a weight decay of 0.01. For fair comparison, we keep the same bias-only model for all the ensemble-based baselines. To tackle the high performance variance on test sets as observed by \citet{clark-etal-2019-dont}, we run each experiment five times and report the mean accuracy scores. 

As for the autoencoder, our multiple experiments reveal that when make sure the bottleneck architecture, the detailed dimension of each layer makes few differences. More implementation details such as $\lambda$, $d$ selection can be found in Section~\ref{sec:analysis}.

\subsection{Experimental Results}

The extensive results of all the above mentioned methods are summarized in Table~\ref{tab: main_res}. The results on the original development and test sets of each task represent the in-distribution performance. Obviously, for all three tasks, RISK improves BERT-base by a large margin on the challenging test set. Moreover, it surpasses other baselines not only for the out-of-distribution test set, but also the in-distribution ones.

\begin{figure}[H]
    \centering
    \includegraphics[scale=0.23]{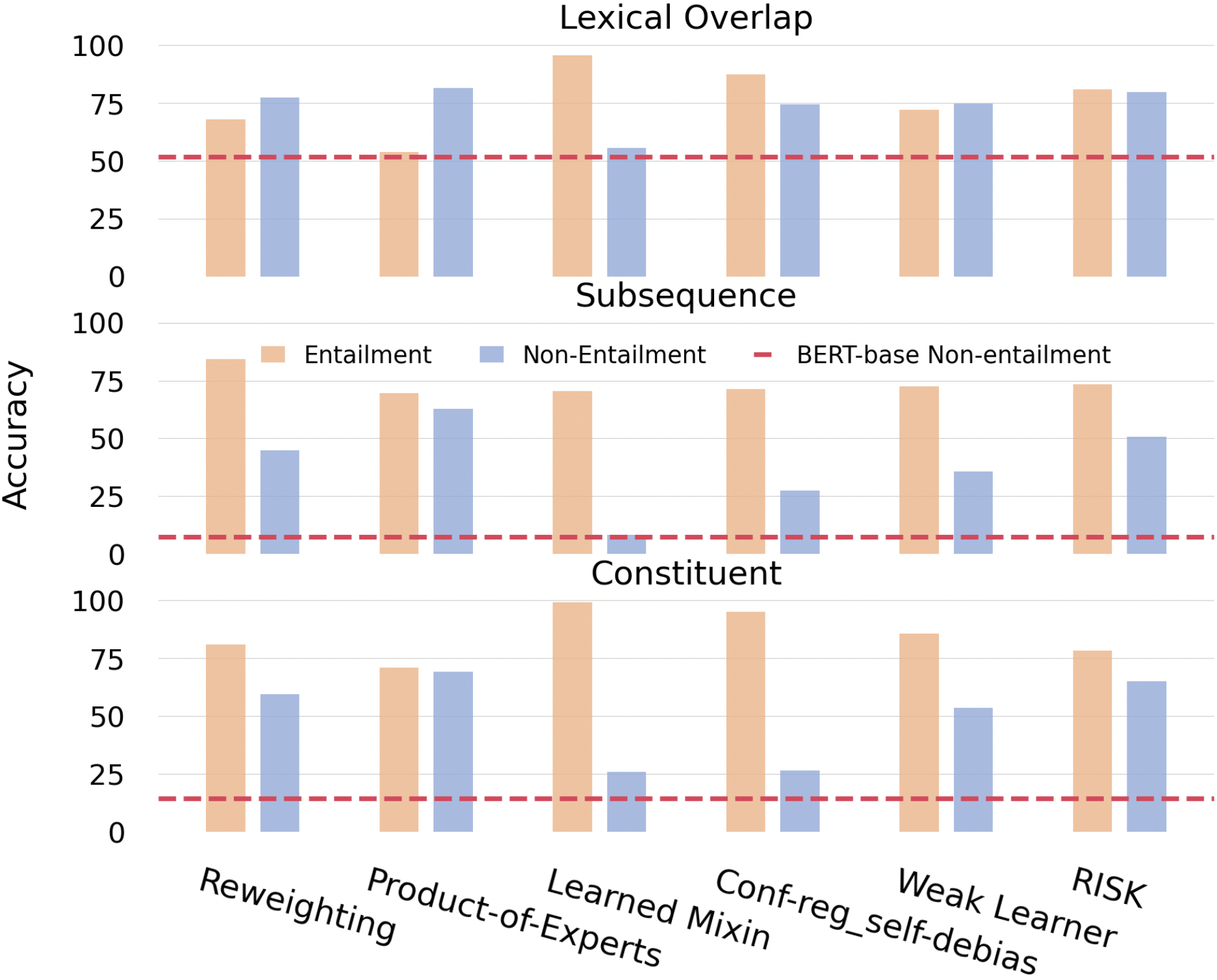}
    \caption{Performance of \textit{RISK} and other baselines on the \textit{entailment} and \textit{non-entailment} categories for each heuristic(i.e., lexical overlap, subsequence and constituent) that HANS was designed to capture.}
    \label{fig:test_hans}
\end{figure}

\textbf{Out-of-distribution generalization and biases mitigation.} The absence of explicit knowledge on bias attributes seemingly create a gap between the generalization ability of bias-known models and bias-unknown models. Though RISK furthur eliminate any supervision of specific bias signal, it still generalize well to the out-of-the distribution. To validate the effectiveness of RISK in mitigating bias, in Figure~\ref{fig:test_hans}, we break down the results on HANS into three different heuristics that the dataset was built upon. The increase of the accuracy in comparison with BERT-base on the \textit{non-entailment} category can reflect the degree to which this bais is removed. Although the overall accuracy of Conf-reg$_{\rm \mathbf{self-debias}}$ on HANS is higher than that of Product-of-experts, as shown in Figure~\ref{fig:test_hans}, it's debiasing capacity is actually the worst. However, RISK can do well in mitigating the three known biases, and is on par with Product-of-Experts, outperforming other baselines.

\textbf{In-distribution performance retention.} The mitigation of dataset bias often suffers from the trade-off between removing shortcut features and sacrificing in-distribution performance. Especially, on PAWS dataset, this trade-off becomes more pronounced. We can observe that previous methods all have a drop in in-distribution test set for MNLI and QQP, which can be attributable to their explicit omission of biased examples. In contrast, our method finds a balance point via intended-feature subspace, where the out-of-distribution performance is improved and the in-distribution is almost retained. For Fever, the in-distribution accuracy of RISK even increases compared to that of BERT-base. 

\textbf{Ablation Studies.} We assumed the reconstruction loss and projection loss are integral parts of RISK as they ensured the intended-feature subspace is \textit{informative} and \textit{shared}. To have an understanding of their impacts on the final performance respectively, we do the ablation studies, and results are shown in Table \ref{tab: main_res}. Comparing the performance degradation, we can conclude that the projection loss plays a key role in helping mitigating dataset bias, and reconstruction loss can be viewed as a regularization that further bound the subspace to be more task-relevant to enhance the accuracy. As can be seen that faced with the removal of reconstruction loss or projection loss, in-distribution performances of the three tasks remain little affected.

\section{Analysis and Discussion}
In this section, we construct supplementary experiments to further analyze RISK's effectiveness. Free of supervision on bias, we reveal that RISK can deal with more challenging scenarios.

\subsection{Hyper-parameter Exploration}\label{sec:analysis}
To recover the intended feature, we introduce two hyperparameters, the weight $\lambda$ of projection loss and the subspace dimension $d$. During the grid search for a fine-grained tuning, we find the values of this two hyperparameters have a close connection with intrinsic properties of dataset.

\paragraph{\textbf{(1). $\lambda$ reflects the hardness of challenging set.}}

In the process of optimizing $\lambda$, we observe that for the three tasks, RISK achieves best out-of-distribution performance with different value of $\lambda$.
For the sake of having a qualitative understanding on the three out-of-distribution test set, we compare the average of sentence length and constituency parse tree height of example in HANS, Fever-Symmetric and PAWS respectively. 

\begin{figure}[h]
    \centering
    \includegraphics[scale=0.18]{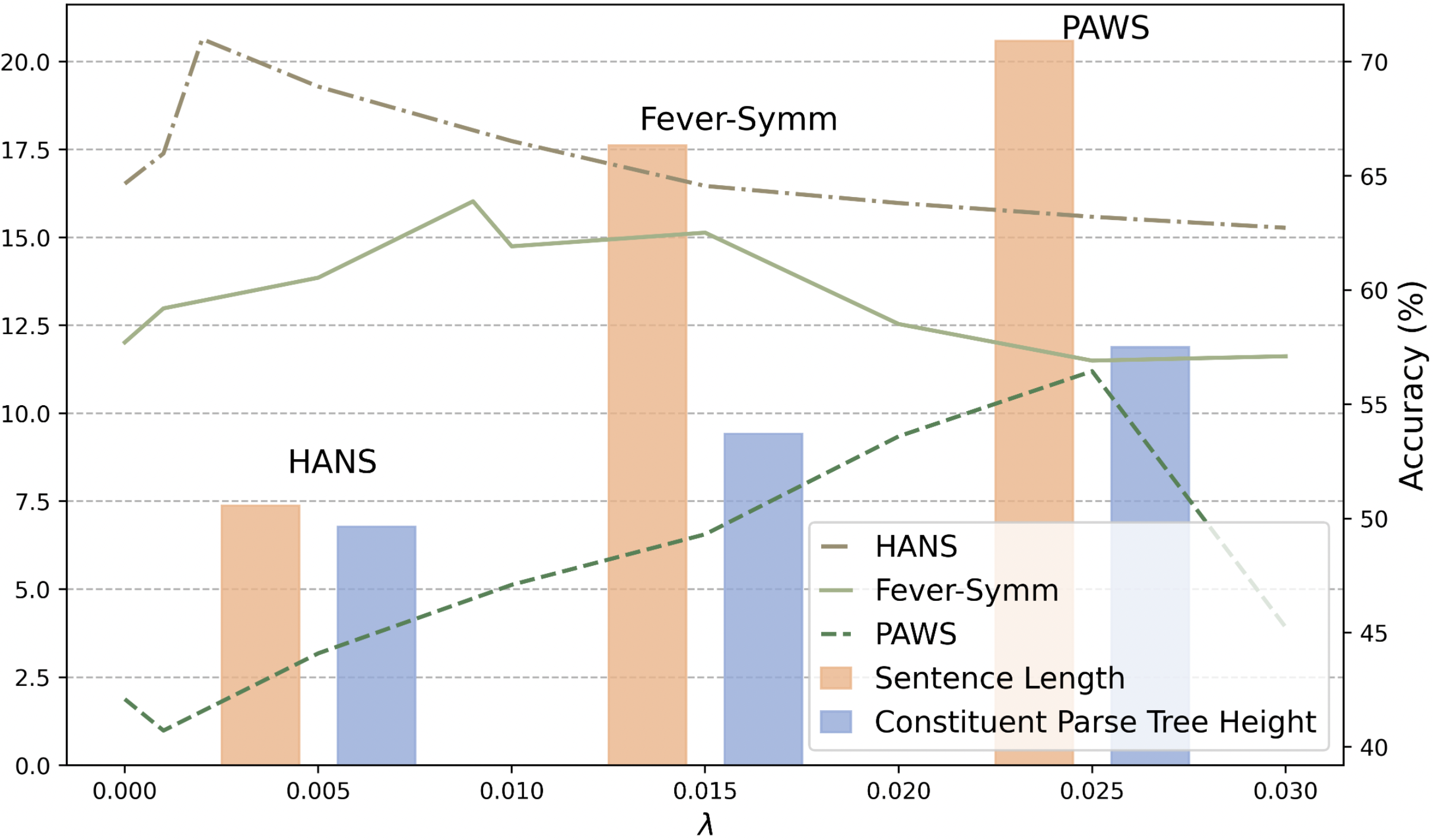}
    \caption{Bar chart represents average sentence length and constituent parse tree height of three out-of-distribution set. Line graph plots model performance with different $\lambda$.}
    \label{fig:lambda_effect}
\end{figure}

As shown in Figure~\ref{fig:lambda_effect}, we can observe that PAWS contains longer and syntactically more complex sentences. In contrast, HANS appears to be more easier for model to learn. Accordingly, easier HANS dataset requires a smaller weight of projection loss to obtain the best performance while PAWS requires a larger $\lambda$ of 0.025. What's more, as for the harder patterns in PAWS for model to generalize,  model performance on this task is more \textit{sensitive} to the change of $\lambda$ in a small range.

\textbf{(2). $d$ reflects the degree of distribution shift.}
 \begin{figure}[h]
    \centering
    \includegraphics[scale=0.18]{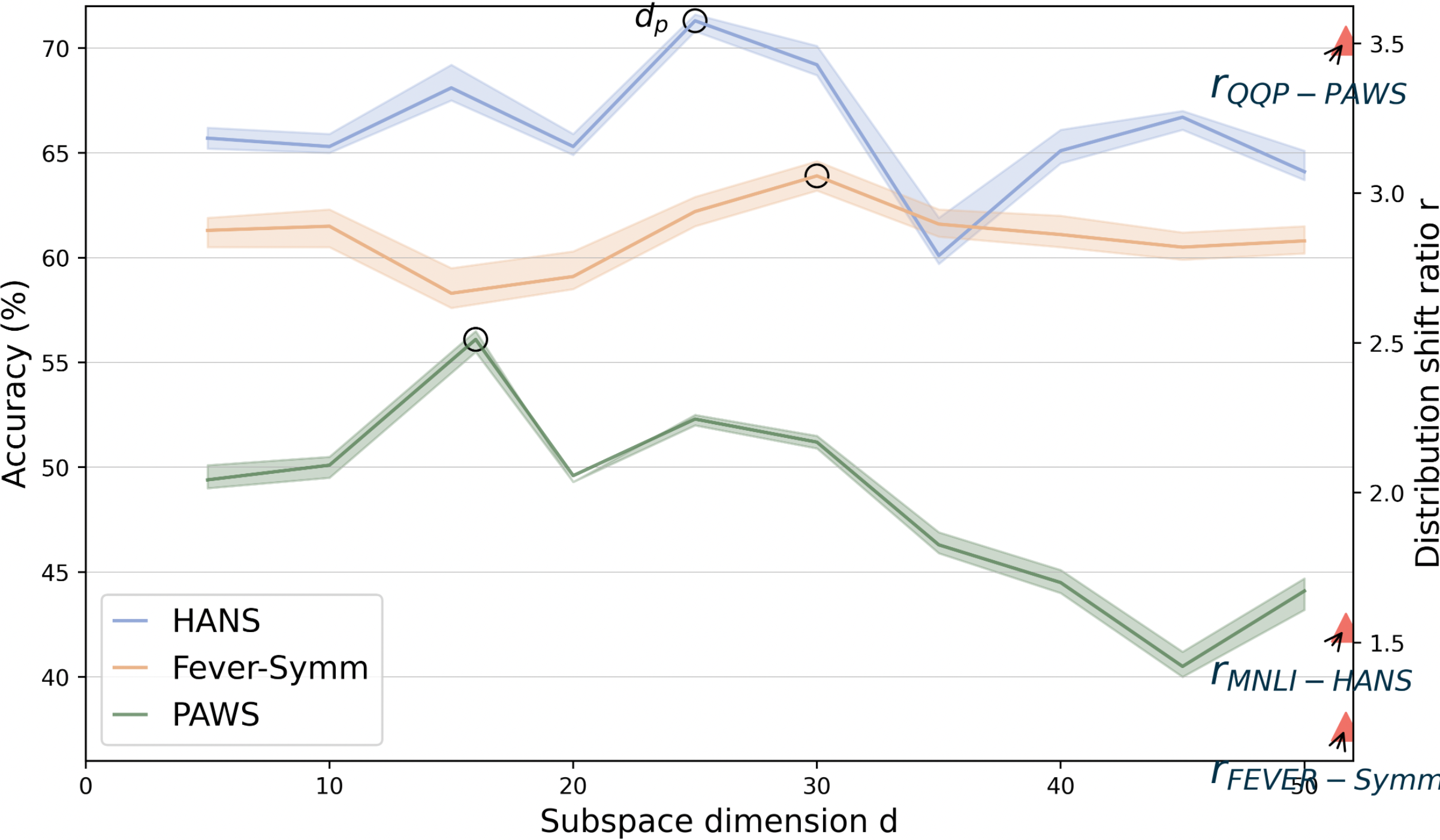}
    \caption{Model performance on HANS, Fever-Symm and PAWS with different subspace dimension $d$.}
    \label{fig:dimension_effect}
\end{figure}

To quantitatively describe the distribution shift, we propose \textit{bias skewness} as an indicator of how biased a dataset is:
\begin{equation}
    {\rm bias\ skewness} = \frac{\#\rm{\ biased\ examples}}{\#\rm{\ bias-free\ examples}} \nonumber
\end{equation}

Thus, the ratio $r$ of bias skewness between ID and OOD can mirror the distribution shift, the larger $r$, the greater distribution discrepancies.
As shown in Figure~\ref{fig:dimension_effect}, denote $d_p$ as the optimal subspace dimension original training data set recovered to peak performance on the out-of-distribution set, and it turns out that ${\rm PAWS}_{d_p} < {\rm HANS}_{d_p} < {\rm Fever\mbox{-}Symm}_{d_p}$. However, the ratio $r$ reflects that $r_{_{\rm QQP-PAWS}} > r_{_{\rm MNLI-HANS}} > r_{_{\rm Fever-Symm}}$.

We can conclude that when faced with a larger distribution shift, the subspace dimension $d$ on in-distribution training set should be smaller. In essence, $d$ can be established a close connection with \textbf{intrinsic dimension}~\cite{ansuini2019intrinsic}, i.e., the minimal number of parameters needed to represent a dataset. As our experiments reveal that a \textbf{16}-dimensional subspace with intended-features can represent the highly biased QQP training dataset well.


\subsection{Transferability Analysis}

We further examine the robustness of our approach along with other baselines by transferring to a more challenging scenario, training on MNLI but testing on Adversarial NLI. In our setting, Adversarial NLI contains not only human-crafted adversarial examples~\cite{nie-etal-2020-adversarial} but also those generated by textual adversarial attacks (TextFooler, \citet{Jin_Jin_Zhou_Szolovits_2020}). In general, models utilizing bias patterns that lack the ability to understand the underlying semantics are vulnerable to be attacked. Results are summarized in Table~\ref{tab:anli} as follows. 
\begin{table}[!htbp]
    \centering
    \scalebox{0.8}{\begin{tabular}{c|ccc|c}
    \hline
       Model  & R1 & R2 & R3 & ANLI-m \\ \hline
       BERT-base  & 0 & 28.9 & 28.8 & 33.0  \\
       Product-of-Experts & \textbf{25.2} & 27.5 & 31.3 & 53.8\\
       Learned-Mixin  & 23.6 & 28.0 & 30.9 & 54.9 \\
       Conf-reg$_{\rm \mathbf{self-debias}}$ & 21.8 & 27.4 & 31.0 & 48.5\\
       RISK  & 25.1 & \textbf{31.2} & \textbf{31.9} & \textbf{57.1} \\
    \hline
    \end{tabular}}
    \caption{Model performance(accu.(\%)) on adversarial MNLI. ANLI R1-R3 are challenging instances designed by human edition on input text. ANLI-m is adversarial MNLI-matched dataset generated by TextFooler based on blackbox BERT. }
    \label{tab:anli}
\end{table}

We can observe that vanilla BERT-base model trained on MNLI are vulnerable to those adversarial examples, especially ones generated by human edition, suggesting BERT relies overly on bias features to make predictions. On the other hand, either bias-known or bias-unknown models can more or less defend against these attacks. Compared to these baselines, RISK can consistently improve performance on all the adversarial test sets. This indicates the intended subspace has the power to robustify NLU models to various distribution shifts.

\section{Related Work}

 We categorize the multiple lines of research devoted to mitigating dataset bias into three paradigms, in accordance with how the supervision is applied for bias mitigation.

\subsection{Supervision from Bias Annotations}
Concerns on robustness give rise to the discovery of a wide variety of biases in existing popular datasets, e.g., models make predictions only rely on the hypothesis in NLI datasets~\cite{Gururangan2018AnnotationAI}. 
\citet{belinkov-etal-2019-adversarial} utilize adversarial training to remove the known hypothesis-only features from model internal representations.
Moreover, the understanding of specific dataset bias motivates the emergence of ensemble-based debiasing methods~\cite{clark-etal-2019-dont,he-etal-2019-unlearn,aclUtamaMG20} , which have shown promising improvements on the out-of-distribution performance. Generally, they view the known dataset biases as prior knowledge and design a simple bias-tailored model, namely the \textit{bias-only model} and factor bias out of the \textit{main model} through ensemble-based training. However, \citet{xiong2021uncertainty} theoretically prove that the inaccurate uncertainty estimations of the bias-only model can hurt the debiasing performance, and they propose to conduct calibration on the bias-only model.

\subsection{Supervision from Model and Training}
The excessive reliance on the assumption that specific types of biased features are known a-prior limits model's transferability. Correspondingly, this line of work seeks for the automatic identification of potentially biased examples, as their empirical results manifest that models with limited capacity~\cite{clark-etal-2020-learning,sanh2021learning} or training on a fewer thousand examples~\cite{utama-etal-2020-towards} exhibit different learning dynamics, and thus can be used to capture relatively shallow correlations.

Meanwhile, other observations have been made that a better use of minority examples(e.g., examples that are under-represented in the training distribution, or examples that are harder to learn) can play role in models' generalization as well.
As \citet{SagawaRKL20} point out, the fundamental reason leading to poor generalization lies in models' behaviour of \textit{memorizing} the minority samples. Particularly,
\citet{Tu2020AnES} leverage the auxiliary tasks to help improve the generalization capability of pre-trained models on the minority groups.  \citet{yaghoobzadeh-etal-2021-increasing} propose to use \textit{example forgetting} to find minority examples and make a second fine-tuning on those minorities. 

\subsection{Supervision from Augmentated Data}

Data augmentation techniques have shown to be effective in regularizing models from overfitting to the training data\citep{novak2018sensitivity}.  In this sense, when distribution shifts, the model will rely little on spurious correlations as a wider variety of predictive features are captured. 
This has attracted interest as a way to remove biases by explicitly modifying the dataset distribution\citep{min-etal-2020-syntactic}.  \citet{Kaushik2020Learning} and \citet{srivastava2020robustness} draw upon human-in-the-loop to augment existing training set with diverse and rich examples of potential unmeasured variables. \citet{Wang2021RobustnessTS} further propose to automatically generate such counterfactual samples via a closet opposite matching strategy.
Different from the augmentation of causal associations between features and classes, \citet{Wang2021IdentifyingAM} apply a cross-data analysis and knowledge-aware perturbations to identify spurious tokens on the stability of model predictions. 


\section{Conclusion}
In this work, we shed light into feature subspace with the aim to create an underlying pathway --- from the biased input examples to robust output prediction. Viewing shortcut features as redundancy, we construct a simple but effective Recovery Layer within the autoencoder structure for bias mitigation. Extensive experiments demonstrate the strengths of our model: better generalization, dataset-agnostic transferability and the robustness to more challenging scenarios. We believe this feature-based debiasing framework opens up new directions for establishing a trustworthy NLU model. Meanwhile, our concise motivation and implementation throw out a thought-provoking question, that is for model, for feature, sometimes less can be better.

\section*{Acknowledgements}
The authors wish to thank the anonymous reviewers for their helpful comments. This work was partially National Natural Science Foundation of China~(No. 62076069, 61976056), Shanghai Municipal Science and Technology Major Project~(No.2021SHZDZX0103), and sponsored by CCF-Tencent Open Fund.


\bibliography{emnlp2021}
\bibliographystyle{acl_natbib}

\appendix

\end{document}